\documentclass[11pt,a4paper]{article}
\usepackage[hyperref]{acl2019}
\usepackage{tikz}
\usepackage{times}
\usepackage{latexsym}
\usepackage{comment}
\usepackage{xspace}
\usepackage{xcolor}
\usepackage{amsmath}
\usepackage{graphicx}
\usepackage{multirow}
\usepackage{booktabs}
\usepackage{amsfonts}
\usepackage{adjustbox}
\usepackage{array}
\usepackage{url}
\usepackage{pifont}

\newcommand{\dataset}{\textit{VIST-Edit}\xspace}

\newcolumntype{x}[1]{>{\centering\arraybackslash}p{#1}}
\newcolumntype{R}[1]{>{\raggedleft\let\newline\\\arraybackslash\hspace{0pt}}m{#1}}
\newcolumntype{L}[1]{>{\raggedright\let\newline\\\arraybackslash\hspace{0pt}}m{#1}}

\aclfinalcopy 

\title{Visual Story Post-Editing}

\author{Ting-Yao Hsu$^1$,~Chieh-Yang Huang$^1$,~Yen-Chia Hsu$^2$,~Ting-Hao (Kenneth) Huang$^1$ \\
  $^1$Pennsylvania State University, State College, PA, USA\\
  $^2$Carnegie Mellon University, Pittsburgh, PA, USA\\
  $^1$\texttt{\{txh357,~chiehyang,~txh710\}@psu.edu}\\
  $^2$\texttt{yenchiah@andrew.cmu.edu}
  }

\date{}

\begin{document}
\maketitle
\begin{abstract}

We introduce the first dataset for \textbf{human edits of machine-generated visual stories} and explore how these collected edits may be used for the \textit{visual story post-editing} task.
The dataset, \dataset\footnote{\dataset: https://github.com/tingyaohsu/VIST-Edit}, includes 14,905 human-edited versions of 2,981 machine-generated visual stories. The stories were generated by two state-of-the-art visual storytelling models, each aligned to 5 human-edited versions.
We establish baselines for the task, showing how a relatively small set of human edits can be leveraged to boost the performance of large visual storytelling models.
We also discuss the weak correlation between automatic evaluation scores and human ratings, motivating the need for new automatic metrics.

\end{abstract}

\section{Introduction}

Professional writers emphasize the importance of \textit{editing}. Stephen King once put it this way: ``to write is human, to \textit{edit} is divine.''~\cite{king2000writing} Mark Twain had another quote: ``Writing is easy. All you have to do is cross out the wrong words.''~\cite{twain1876adventures}
Given that professionals revise and rewrite their drafts intensively, machines that generate stories may also benefit from a good editor.
Per the evaluation of the first Visual Storytelling Challenge~\cite{mitchell2018proceedings}, the ability of an algorithm to tell a sound story is still far from that of a human. Users will inevitably need to edit generated stories before putting them to real uses, such as sharing on social media.

\begin{figure}[t]
    \centering
    \includegraphics[width=1.0\columnwidth]{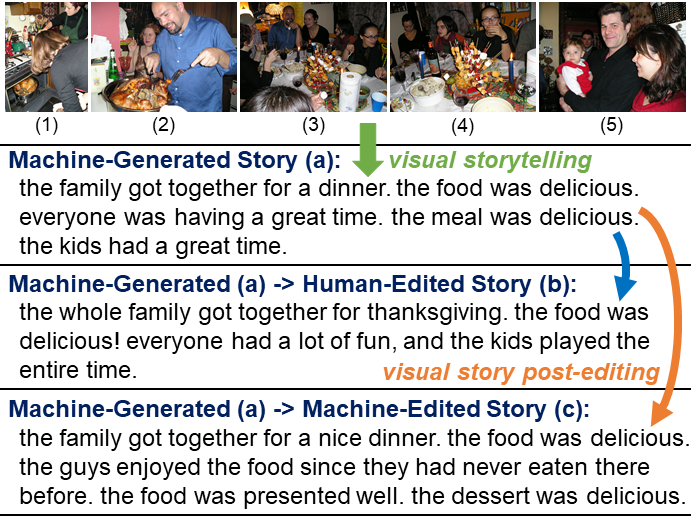}
    \caption{A machine-generated visual story (a) (by GLAC), its human-edited (b) and machine-edited (c) (by LSTM) version.}
    \label{fig:vist-ape-example}
\end{figure}


We introduce the first dataset for \textbf{human edits of machine-generated visual stories}, \dataset, and explore how these collected edits may be used for the task of \textit{visual story post-editing} (see Figure~\ref{fig:vist-ape-example}).
The original visual storytelling (VIST) task, as introduced by Huang {\em et al.}~\shortcite{huang2016visual}, takes a sequence of five photos as input and generates a short story describing the photo sequence.
Huang {\em et al.} also released the VIST dataset, containing 20,211 photo sequences, aligned to human-written stories.
On the other hand, the automatic post-editing task revises the story generated from visual storytelling models, given both a machine-generated story and a photo sequence. 
Automatic post-editing treats the VIST system as a black box that is fixed and not modifiable.
Its goal is to correct systematic errors of the VIST system and leverage the user edit data to improve story quality.

In this paper, we \textit{(i)} collect human edits for machine-generated stories from two different state-of-the-art models, \textit{(ii)} analyze what people edited, and \textit{(iii)} advance the task of visual story post-editing.
In addition, we establish baselines for the task, and discuss the weak correlation between automatic evaluation scores and human ratings, motivating the need for new metrics.

\section{Related Work}

The visual story post-editing task is related to {\em (i)} automatic post-editing and {\em (ii)} stylized visual captioning. \textbf{Automatic post-editing (APE)} revises the text generated typically from a machine translation (MT) system, given both the source sentences and translated sentences. Like the proposed VIST post-editing task, APE aims to correct the systematic errors of MT, reducing translator workloads and increasing productivity~\cite{astudillo2018proceedings}. 
Recently, neural models have been applied to APE in a sentence-to-sentence manner~\cite{libovicky2016cuni,junczys2016log}, differing from previous phrase-based models that translate and reorder phrase segments for each sentence, such as~\cite{simard2007statistical,bechara2011statistical}. More sophisticated sequence-to-sequence models with the attention mechanism were also introduced~\cite{junczys2017exploration,libovicky2017attention}. While this line of work is relevant and encouraging, it has not explored much in a creative writing context.
It is noteworthy that Roemmele {\em et al.} previously developed an online system, Creative Help, for collecting human edits for computer-generated narrative text~\cite{roemmele2018automated}.
The collected data could be useful for story APE tasks.

Visual story post-editing could also be considered relevant to \textbf{style transfer on image captions}. Both tasks take images and source text ({\em i.e.}, machine-generated stories or descriptive captions) as inputs and generate modified text ({\em i.e.}, post-edited stories or stylized captions). End-to-end neural models have been applied to the transfer styles of image captions. For example, \textit{StyleNet}, an encoder-decoder-based model trained on paired images and factual captions together with an unlabeled stylized text corpus, can transfer descriptive image captions to creative captions, {\em e.g.,} humorous or romantic~\cite{gan2017styleNet}. Its advanced version with an attention mechanism, \textit{SemStyle}, was also introduced~\cite{mathews2018semstyle}.
In this paper, we adopt the APE approach to treat pre- and post-edited stories as parallel data instead of the style transfer approach that omits this parallel relationship during model training.

\section{Dataset Construction \& Analysis}
\paragraph{Obtaining Machine-Generated Visual Stories} 
This \dataset dataset contains visual stories generated by two state-of-the-art models, \textbf{GLAC} and \textbf{AREL}.
GLAC (Global-Local Attention Cascading Networks)~\cite{kim2018glac} achieved the highest human evaluation score in the first VIST Challenge~\cite{mitchell2018proceedings}.
We obtain the pre-trained GLAC model provided by the authors via Github and run it on the entire VIST test set and obtain 2,019 stories.
AREL (Adversarial REward Learning)~\cite{acl2018wang} was the earliest available implementation online, and achieved the highest METEOR score on public test set in the VIST Challenge.
We also acquire a small set of human edits for 962 AREL's stories generated using VIST test set, collected by Hsu {\em et al.}~\shortcite{hsu2019users}.

\begin{figure}[t]
    \centering
    \includegraphics[width=1.0\columnwidth]{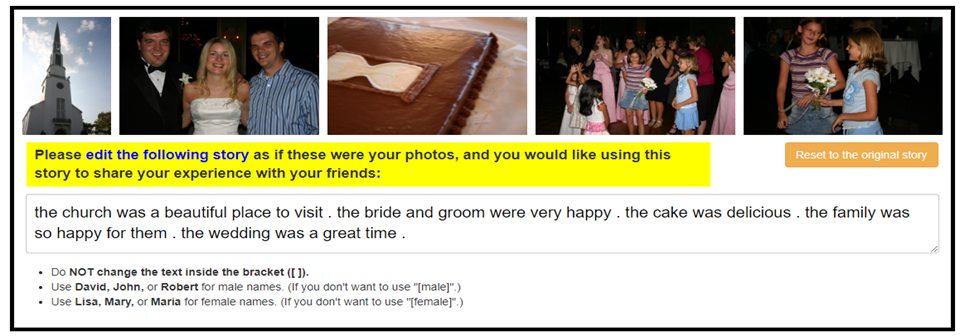}
    \caption{Interface for visual story post-editing. An instruction (not shown to save space) is given and workers are asked to stick with the plot of the original story.}
    \label{fig:ui}
\end{figure}

\paragraph{Crowdsourcing Edits}
For each machine-generated visual story, we recruit five crowd workers from Amazon Mechanical Turk (MTurk) to revise it (at \$0.12/HIT,) respectively.
We instruct workers to edit the story ``as if these were your photos, and you would like using this story to share your experience with your friends.''
We also ask workers to stick with the photos of the original story so that workers would not ignore the machine-generated story and write a new one from scratch.
Figure~\ref{fig:ui} shows the interface.
For GLAC, we collect 2,019 $\times$ 5 = 10,095 edited stories in total; and for AREL, 962 $\times$ 5 = 4,810 edited stories have been collected by Hsu {\em et al.}~\shortcite{hsu2019users}.

\paragraph{Data Post-processing}
We tokenize all stories using CoreNLP~\cite{coreNLP} and replace all people names with generic [male/female] tokens.
Each of GLAC and AREL set is released as training, validation, and test following an 80\%, 10\%, 10\% split, respectively.

\subsection{What do people edit?}
\begin{figure}[thbp]
    \centering
    \includegraphics[width=1.0\columnwidth]{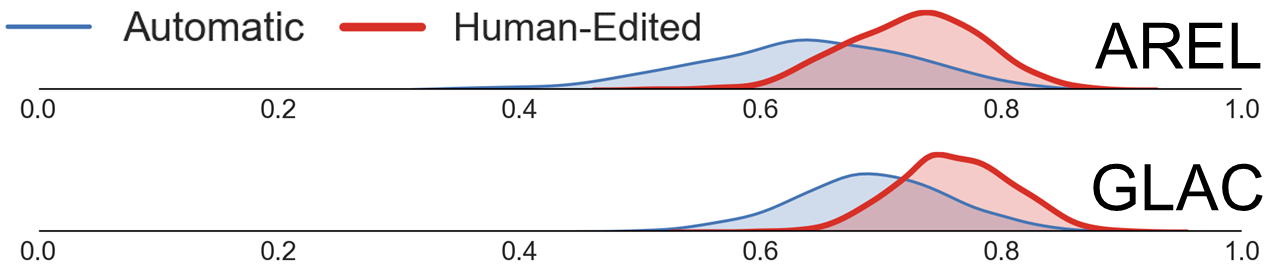}
    \caption{KDE plot of type-token ratio (TTR) for pre-/post-edited stories. People increase lexical diversity in machine-generated stories for both AREL and GLAC.}
    \label{fig:ttr}
\end{figure}

We analyze human edits for GLAC and AREL.
First, crowd workers systematically \textbf{increase lexical diversity}.
We use type-token ratio (TTR), the ratio between the number of word types and the number of tokens, to estimate the lexical diversity of a story~\cite{1af8cfb45ecf4823b5e14c69b80d4d5a}.
Figure~\ref{fig:ttr} shows significant (p\textless.001, paired t-test) positive shifts of TTR for both AREL and GLAC, which confirms the findings in Hsu {\em et al.}~\shortcite{hsu2019users}.
Figure~\ref{fig:ttr} also indicates that GLAC generates stories with higher lexical diversity than that of AREL.

Second, people \textbf{shorten AREL's stories but lengthen GLAC's stories.}
We calculate the average number of Part-Of-Speech (POS) tags for tokens in each story using the python NLTK~\cite{bird2009natural} package, as shown in Table~\ref{tab:pos-stats}.
We also find that the average number of tokens in an AREL story (43.0, SD=5.0) decreases (41.9, SD=5.6) after human editing, while 
that of GLAC (35.0, SD=4.5) increases (36.7, SD=5.9).
Hsu has observed that people often replace ``determiner/article + noun'' phrases ({\em e.g.,} ``a boy'') with pronouns ({\em e.g.,} ``he'') in AREL stories~\shortcite{hsu2019users}.
However, this observation cannot explain the story lengthening in GLAC, where each story on average has an increased 0.9 nouns after editing.
Given the average per-story edit distances~\cite{levenshtein1966binary,damerau1964technique} for AREL (16.84, SD=5.64) and GLAC (17.99, SD=5.56) are similar, this difference is unlikely to be caused by deviation in editing amount.

\begin{table}[h]
    \scriptsize
    \centering
    \addtolength{\tabcolsep}{-0.165cm}
    \begin{tabular}{lrrrrrrrrrrr}
        \toprule
        \resizebox{0.55cm}{!}{\textbf{AREL}}     &    \textbf{.} & \textbf{ADJ} &  \textbf{ADP} &  \textbf{ADV} & \textbf{CONJ} &  \textbf{DET} & \textbf{NOUN} & \textbf{PRON} & \textbf{PRT} & \textbf{VERB} & \textbf{Total} \\ \hline
        \textbf{Pre}      &  5.2 & 3.1 &  3.5 &  1.9 &  0.5 &  8.1 & 10.1 &  2.1 & 1.6 &  6.9 & 43.0 \\
        \textbf{Post}     &  4.7 & 3.1 &  3.4 &  1.9 &  0.8 &  7.1 &  9.9 &  2.3 & 1.6 &  7.0 & 41.9 \\
        \textbf{$\Delta$} & -0.5 & 0.0 & -0.1 & -0.1 &  0.4 & -1.0 & -0.2 &  0.2 & 0.0 &  0.1 & -1.2 \\ 
        \bottomrule
        \toprule
        \resizebox{0.55cm}{!}{\textbf{GLAC}}     &    \textbf{.} & \textbf{ADJ} &  \textbf{ADP} &  \textbf{ADV} & \textbf{CONJ} &  \textbf{DET} & \textbf{NOUN} & \textbf{PRON} & \textbf{PRT} & \textbf{VERB} & \textbf{Total} \\ \hline
        \textbf{Pre}      &  5.0 &  3.3 & 1.7 &  1.9 &  0.2 &  6.5 &  7.4 &  1.2 & 0.8 &  6.9 & 35.0 \\
        \textbf{Post}     &  4.5 &  3.2 & 2.4 &  1.8 &  0.8 &  6.1 &  8.3 &  1.5 & 1.0 &  7.0 & 36.7 \\ 
        \textbf{$\Delta$} & -0.5 & -0.1 & 0.7 & -0.1 &  0.6 & -0.3 &  0.9 &  0.3 & 0.2 &  0.1 &  1.7 \\ 
        \bottomrule
    \end{tabular}
    \addtolength{\tabcolsep}{0.165cm}
    \caption{Average number of tokens with each POS tag per story. ($\Delta$: the differences between post- and pre-edit stories. NUM is omitted because it is nearly 0. Numbers are rounded to one decimal place.)}\label{tab:pos-stats}
\end{table}

\textit{Deleting} extra words requires much less time than other editing operations~\cite{popovic2014relations}.
Per Figure~\ref{fig:ttr}, AREL's stories are much more repetitive.
We further analyze the type-token ratio for nouns (${TTR}_{noun}$) and find AREL generates duplicate nouns.
The average ${TTR}_{noun}$ of an AREL's story is 0.76 while that of GLAC is 0.90.
For reference, the average ${TTR}_{noun}$ of a human-written story (the entire VIST dataset) is 0.86.
Thus, we hypothesize workers prioritized their efforts in deleting repetitive words for AREL, resulting in the reduction of story length.

\section{Baseline Experiments}

\begin{table*}[t]
\centering
\scriptsize	
\addtolength{\tabcolsep}{-0.03cm} 
\begin{tabular}{|l|rrrrrr|rrrrrr|}
\hline
\multicolumn{1}{|c|}{\textit{}} & \multicolumn{6}{c|}{\textit{\textbf{AREL}}} & \multicolumn{6}{c|}{\textit{\textbf{GLAC}}} \\ \hline
Edited By & Focus & Coherence & Share & Human & Grounded & Detailed & Focus & Coherence & Share & Human & Grounded & Detailed \\ \hline
\textbf{N/A} & 3.487 & 3.751 & 3.763 & 3.746 & 3.602 & 3.761 & 3.878 & 3.908 & 3.930 & 3.817 & 3.864 & 3.938 \\ \hline
\textbf{TF (T)} & 3.433 & 3.705 & 3.641 & 3.656 & 3.619 & 3.631 & 3.717 & 3.773 & 3.863 & 3.672 & 3.765 & 3.795 \\
\textbf{TF (T+I)} & \textbf{3.542} & 3.693 & 3.676 & 3.643 & 3.548 & 3.672 & 3.734 & 3.759 & 3.786 & 3.622 & 3.758 & 3.744 \\
\textbf{LSTM (T)} & \textbf{3.551} & \textbf{3.800} & \textbf{3.771} & \textbf{3.751} & \textbf{3.631} & \textbf{3.810} & \textbf{3.894} & 3.896 & 3.864 & \textbf{3.848} & 3.751 & 3.897 \\
\textbf{LSTM (T+I)} & \textbf{3.497} & 3.734 & 3.746 & 3.742 & 3.573 & 3.755 & 3.815 & 3.872 & 3.847 & 3.813 & 3.750 & 3.869 \\ \hline
\textbf{Human} & 3.592 & 3.870 & 3.856 & 3.885 & 3.779 & 3.878 & 4.003 & 4.057 & 4.072 & 3.976 & 3.994 & 4.068 \\ \hline
\end{tabular}
\addtolength{\tabcolsep}{0.03cm} 
\caption{Human evaluation results. Five human judges on MTurk rate each story on the following six aspects, using a 5-point Likert scale (from Strongly Disagree to Strongly Agree):
Focus,
Structure and Coherence, 
Willing-to-Share (``I Would Share''),
Written-by-a-Human (``This story sounds like it was written by a human.''),
Visually-Grounded, and
Detailed. We take the average of the five judgments as the final score for each story.
LSTM(T) improves all aspects for stories by AREL, and improves ``Focus'' and ``Human-like'' aspects for stories by GLAC.
}
\label{tab:human-eval}
\end{table*}

\begin{figure*}[t]
    \centering
    \includegraphics[width=1.0\textwidth]{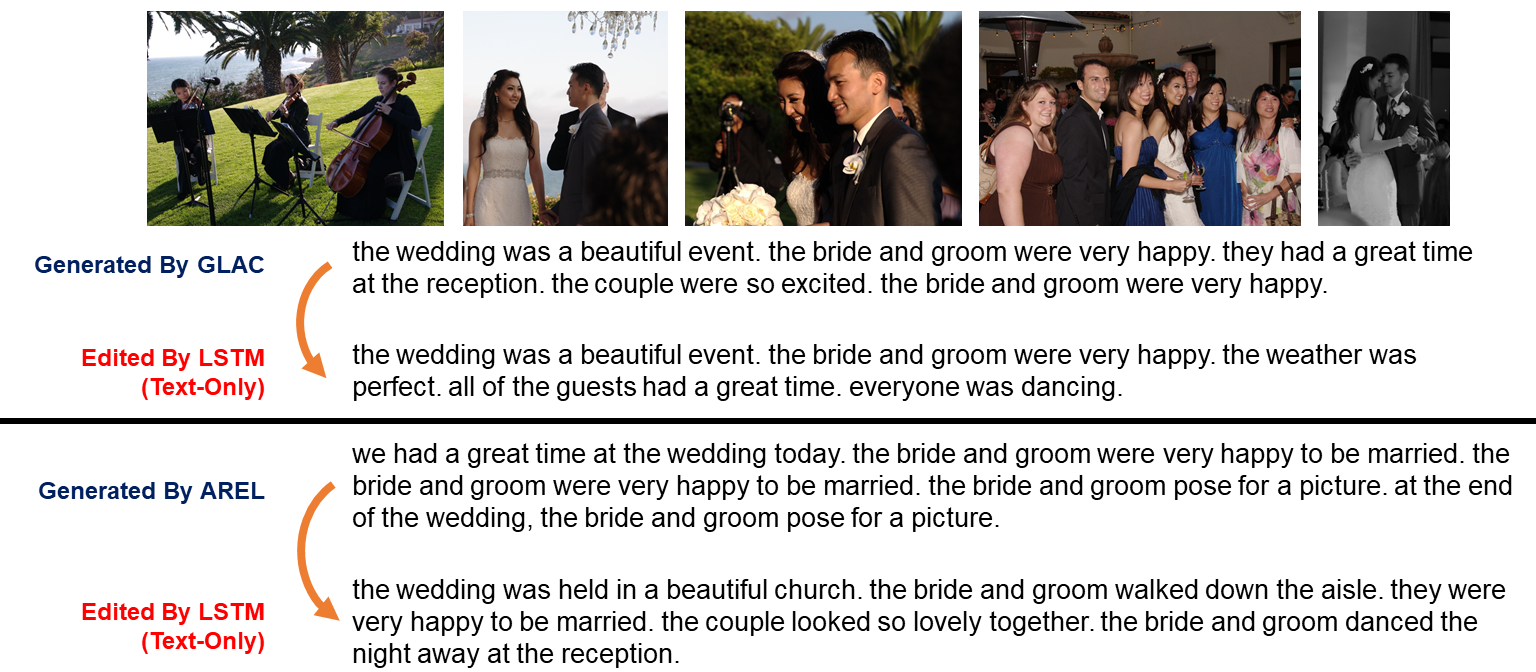}
    \caption{Example stories generated by baselines.}
    \label{fig:representative-story}
\end{figure*}

We report baseline experiments on the visual story post-editing task in Table~\ref{tab:human-eval}.
AREL's post-editing models are trained on the augmented AREL training set and evaluated on the AREL test set of \dataset, and GLAC's models are tested using GLAC sets, too.
Figure~\ref{fig:representative-story} shows examples of the output.
Human evaluations (Table~\ref{tab:human-eval}) indicate that the post-editing model improves visual story quality.

\subsection{Methods}

Two neural approaches, Long short-term memory (LSTM) and Transformer, are used as baselines, where we experiment using {\em (i)} text only (T) and {\em (ii)} both text and images (T+I) as inputs.

\paragraph{LSTM}
An LSTM seq2seq model is used~\cite{sutskever2014sequence}.
For the text-only setting, the original stories and the human-edited stories are treated as source-target pairs.
For the text-image setting, we first extract the image features using the pre-trained ResNet-152 model~\cite{he2016deep} and represent each image as a 2048-dimensional vector. 
We then apply a dense layer on image features in order to both fit its dimension to the word embedding and learn the adjusting transformation. 
By placing the image features in front of the sequence of text embedding, the input sequence becomes a matrix $\in \mathbb{R}^{(5+len) \times dim}$, where $len$ is the text sequence length, $5$ means 5 photos, and $dim$ is the dimension of the word embedding.
The input sequence with both image information and text information is then encoded by LSTM, identical as in the text-only setting.

\paragraph{Transformer (TF)}
We also use the Transformer architecture~\cite{vaswani2017attention} as baseline.
The text-only setup and image feature extraction are identical to that of LSTM.
For Transformer, the image features are attached at the end of the sequence of text embedding to form an image-enriched embedding.
It is noteworthy that the position encoding is only applied on text embedding.
The input matrix $\in \mathbb{R}^{(len+5) \times dim}$ is then passed into the Transformer as in the text-only setting.

\subsection{Experimental Setup and Evaluation}

\paragraph{Data Augmentation}
In order to obtain sufficient training samples for neural models, we pair \textit{less}-edited stories with \textit{more}-edited stories of the same photo sequence to augment the data.
In \dataset, five human-edited stories are collected for each photo sequence.
We use the human-edited stories that are less edited -- measured by its Normalized Damerau-Levenshtein distance \cite{levenshtein1966binary,damerau1964technique} to the original story -- as the source and pair them with the stories that are more edited (as the target.)
This data augmentation strategy gives us in total fifteen ($\left ( ^5_2 \right )+5=15$) training samples given five human-edited stories.

\paragraph{Human Evaluation}
Following the evaluation procedure of the first VIST Challenge~\cite{mitchell2018proceedings}, for each visual story, we recruit five human judges on MTurk to rate it on six aspects (at \$0.1/HIT.)
We take the average of the five judgments as the final scores for the story.
Table~\ref{tab:human-eval} shows the results.
The LSTM using text-only input outperforms all other baselines.
It improves all six aspects for stories by AREL, and improves ``Focus'' and ``Human-like'' aspects for stories by GLAC.
These results demonstrate that a relatively small set of human edits can be used to boost the story quality of an existing large VIST model.
Table~\ref{tab:human-eval} also suggests that the quality of a post-edited story is heavily decided by its pre-edited version.
Even after editing by human editors, AREL's stories still do not achieve the quality of pre-edited stories by GLAC.
The inefficacy of image features and Transformer model might be caused by the small size of \dataset.
It also requires further research to develop a post-editing model in a multimodal context.

\section{Discussion}
\paragraph{Automatic evaluation scores do not reflect the quality improvements.}
APE for MT has been using automatic metrics, such as BLEU, to benchmark progress~\cite{libovicky2016cuni}.
However, classic automatic evaluation metrics fail to capture the signal in human judgments for the proposed visual story post-editing task.
%
We first use the human-edited stories as references, but all the automatic evaluation metrics generate \textit{lower} scores when human judges give a higher rating (Table~\ref{tab:AREL-auto-eval}.)

\begin{table}[h]
\center
\scriptsize
\addtolength{\tabcolsep}{-0.12cm} 
\begin{tabular}{@{}lrrrrr@{}}
 & \multicolumn{4}{c}{Reference: AREL Stories \textbf{Edited} by Human} & \multicolumn{1}{l}{} \\ \toprule
 & \textbf{BLEU4} & \textbf{METEOR} & \textbf{ROUGE} & \textbf{Skip-Thoughts} & \textbf{\begin{tabular}[c]{@{}r@{}}Human\\ Rating\end{tabular}} \\ \midrule
\textbf{AREL} & 0.93 & 0.91 & 0.92 & 0.97 & \textbf{3.69} \\ \midrule
\textbf{\begin{tabular}[c]{@{}l@{}}AREL Edited\\ By LSTM(T)\end{tabular}} & 0.21 & 0.46 & 0.40 & 0.76 & \textbf{3.81} \\ \bottomrule
\end{tabular}
\addtolength{\tabcolsep}{0.12cm} 
\caption{Average evaluation scores for AREL stories, using the human-edited stories as references. All the automatic evaluation metrics generate lower scores when human judges give a higher rating.}
\label{tab:AREL-auto-eval}
\end{table}

We then switch to use the human-written stories (VIST test set) as references, but again, all the automatic evaluation metrics generate \textit{lower} scores even when the editing was done by human (Table~\ref{tab:GLAC-auto-eval}.)

\begin{table}[h]
\center
\scriptsize
\begin{tabular}{@{}lrrrr@{}}
 & \multicolumn{4}{c}{Reference: Human-\textbf{Written} Stories} \\ \toprule
 & \textbf{BLEU4} & \textbf{METEOR} & \textbf{ROUGE} & \textbf{Skip-Thoughts} \\ \midrule
\textbf{GLAC} & 0.03 & 0.30 & 0.26 & 0.66 \\ \midrule
\textbf{\begin{tabular}[c]{@{}l@{}}GLAC Edited\\ By Human\end{tabular}} & 0.02 & 0.28 & 0.24 & 0.65 \\ \bottomrule
\end{tabular}
\caption{Average evaluation scores on GLAC stories, using human-written stories as references. All the automatic evaluation metrics generate lower scores even when the editing was done by human.}
\label{tab:GLAC-auto-eval}
\end{table}

\begin{table}[t]
\centering
\scriptsize
\addtolength{\tabcolsep}{-0.13cm} 
\begin{tabular}{llrrrr}
 & & \multicolumn{4}{c}{Spearman rank-order correlation $\rho$} \\
\toprule \hline
  & \textbf{Data Includes}  & \textbf{BLEU4} & \textbf{METEOR} & \textbf{ROUGE} & \textbf{Skip-Thoughts}\\ \hline
\ding{172} & AREL       & .110   & .099  & .063  & .062           \\
\ding{173} &  LSTM-Edited AREL & .106   & .109  & .067 & .205         \\
\ding{174} & \ding{172}+\ding{173}                     & .095   & .092  & .059    & .116              \\ \hline
\ding{175} & GLAC   & .222   & .203  & .140    &  .151             \\
\ding{176} &  LSTM-Edited GLAC                & .163   & .176  & .138    &    .087           \\
\ding{177} & \ding{175}+\ding{176}                     & .196   & .194  & .148    & .116              \\ \hline
\ding{178} & \ding{172}+\ding{175}                     & .091   & .086  & .059    & .088              \\
\ding{179} & \ding{173}+\ding{176}                     & .089   & .103  & .067    & .101              \\
\ding{180} & \ding{172}+\ding{173}+\ding{175}+\ding{176} & .090   & .096  & .069 & .094             \\ \hline \bottomrule
\end{tabular}
\addtolength{\tabcolsep}{0.13cm} 
\caption{Spearman rank-order correlation $\rho$ between the automatic evaluation scores (sum of all six aspects) and human judgment. 
When comparing
among machine-edited stories (\ding{173} and \ding{176}),
among pre- and post-edited stories (\ding{174} and \ding{177}),
or among any combinations of them (\ding{178}, \ding{179} and \ding{180}),
all metrics result in weak correlations with human judgments.
}
\label{tab:auto-human-cor}
\end{table}

Table~\ref{tab:auto-human-cor} further shows the Spearman rank-order correlation $\rho$ between the automatic evaluation scores (sum of all six aspects) and human judgment calculated using different data combination.
In row~\ding{175} of Table~\ref{tab:auto-human-cor}, the reported correlation $\rho$ of METEOR is consistent with the findings in Huang {\em et al.}~\shortcite{huang2016visual}, which suggests that METEOR could be useful when comparing among stories generated by the same visual storytelling model.
However, when comparing
among machine-edited stories (row \ding{173} and \ding{176}),
among pre- and post-edited stories (row \ding{174} and \ding{177}),
or among any combinations of them (row \ding{178}, \ding{179} and \ding{180}),
all metrics result in weak correlations with human judgments.
%
These results strongly suggest the need of a new automatic evaluation metric for visual story post-editing task.
Some new metrics have recently been introduced using linguistic~\cite{roemmele2018linguistic} or story features~\cite{purdy2018predicting} to evaluate story automatically.
More research is needed to examine whether these metrics are useful for story post-editing tasks too.

\section{Conclusion}
\dataset, the first dataset for \textbf{human edits of machine-generated visual stories}, is introduced.
We argue that human editing on machine-generated stories is unavoidable, and such edited data can be leveraged to enable automatic post-editing.
We have established baselines for the task of visual story post-editing, and have motivated the need for a new automatic evaluation metric.

\bibliography{acl2019}

\begin{thebibliography}{28}
\expandafter\ifx\csname natexlab\endcsname\relax\def\natexlab#1{#1}\fi

\bibitem[{Astudillo et~al.(2018)Astudillo, Gra{\c{c}}a, and
  Martins}]{astudillo2018proceedings}
Ram{\'o}n Astudillo, Jo{\~a}o Gra{\c{c}}a, and Andr{\'e} Martins. 2018.
\newblock Proceedings of the amta 2018 workshop on translation quality
  estimation and automatic post-editing.
\newblock In \emph{Proceedings of the AMTA 2018 Workshop on Translation Quality
  Estimation and Automatic Post-Editing}.

\bibitem[{B{\'e}chara et~al.(2011)B{\'e}chara, Ma, and van
  Genabith}]{bechara2011statistical}
Hanna B{\'e}chara, Yanjun Ma, and Josef van Genabith. 2011.
\newblock Statistical post-editing for a statistical mt system.
\newblock In \emph{MT Summit}, volume~13.

\bibitem[{Bird et~al.(2009)Bird, Klein, and Loper}]{bird2009natural}
Steven Bird, Ewan Klein, and Edward Loper. 2009.
\newblock \emph{Natural language processing with Python: analyzing text with
  the natural language toolkit}.
\newblock O'Reilly Media, Inc.

\bibitem[{Damerau(1964)}]{damerau1964technique}
Fred~J Damerau. 1964.
\newblock A technique for computer detection and correction of spelling errors.
\newblock \emph{Communications of the ACM}, 7(3):171--176.

\bibitem[{Gan et~al.(2017)Gan, Gan, He, Gao, and Deng}]{gan2017styleNet}
C.~Gan, Z.~Gan, X.~He, J.~Gao, and L.~Deng. 2017.
\newblock \href {https://doi.org/10.1109/CVPR.2017.108} {Stylenet: Generating
  attractive visual captions with styles}.
\newblock In \emph{2017 IEEE Conference on Computer Vision and Pattern
  Recognition (CVPR)}, pages 955--964.

\bibitem[{Hardie and McEnery(2006)}]{1af8cfb45ecf4823b5e14c69b80d4d5a}
Andrew Hardie and Tony McEnery. 2006.
\newblock \emph{Statistics.}, volume~12, pages 138--146. Elsevier.

\bibitem[{He et~al.(2016)He, Zhang, Ren, and Sun}]{he2016deep}
Kaiming He, Xiangyu Zhang, Shaoqing Ren, and Jian Sun. 2016.
\newblock Deep residual learning for image recognition.
\newblock In \emph{Proceedings of the IEEE conference on computer vision and
  pattern recognition}, pages 770--778.

\bibitem[{Hsu et~al.(2019)Hsu, Hsu, and Huang}]{hsu2019users}
Ting-Yao Hsu, Yen-Chia Hsu, and Ting-Hao~K. Huang. 2019.
\newblock On how users edit computer-generated visual stories.
\newblock In \emph{Proceedings of the 2019 CHI Conference Extended Abstracts
  (Late-Breaking-Work) on Human Factors in Computing Systems}. ACM.

\bibitem[{Huang et~al.(2016)Huang, Ferraro, Mostafazadeh, Misra, Agrawal,
  Devlin, Girshick, He, Kohli, Batra et~al.}]{huang2016visual}
Ting-Hao~Kenneth Huang, Francis Ferraro, Nasrin Mostafazadeh, Ishan Misra,
  Aishwarya Agrawal, Jacob Devlin, Ross Girshick, Xiaodong He, Pushmeet Kohli,
  Dhruv Batra, et~al. 2016.
\newblock Visual storytelling.
\newblock In \emph{Proceedings of the 2016 Conference of the North American
  Chapter of the Association for Computational Linguistics: Human Language
  Technologies}, pages 1233--1239.

\bibitem[{Junczys-Dowmunt and Grundkiewicz(2016)}]{junczys2016log}
Marcin Junczys-Dowmunt and Roman Grundkiewicz. 2016.
\newblock Log-linear combinations of monolingual and bilingual neural machine
  translation models for automatic post-editing.
\newblock \emph{arXiv preprint arXiv:1605.04800}.

\bibitem[{Junczys-Dowmunt and Grundkiewicz(2017)}]{junczys2017exploration}
Marcin Junczys-Dowmunt and Roman Grundkiewicz. 2017.
\newblock An exploration of neural sequence-to-sequence architectures for
  automatic post-editing.
\newblock \emph{arXiv preprint arXiv:1706.04138}.

\bibitem[{Kim et~al.(2018)Kim, Heo, Son, Park, and Zhang}]{kim2018glac}
Taehyeong Kim, Min-Oh Heo, Seonil Son, Kyoung-Wha Park, and Byoung-Tak Zhang.
  2018.
\newblock Glac net: Glocal attention cascading networks for multi-image cued
  story generation.
\newblock \emph{arXiv preprint arXiv:1805.10973}.

\bibitem[{King(2000)}]{king2000writing}
Stephen King. 2000.
\newblock On writing: A memoir ofthe craft.
\newblock \emph{New Yorle: Scrihner}.

\bibitem[{Levenshtein(1966)}]{levenshtein1966binary}
Vladimir~I Levenshtein. 1966.
\newblock Binary codes capable of correcting deletions, insertions, and
  reversals.
\newblock In \emph{Soviet physics doklady}, volume~10, pages 707--710.

\bibitem[{Libovick{\`y} and Helcl(2017)}]{libovicky2017attention}
Jind{\v{r}}ich Libovick{\`y} and Jind{\v{r}}ich Helcl. 2017.
\newblock Attention strategies for multi-source sequence-to-sequence learning.
\newblock \emph{arXiv preprint arXiv:1704.06567}.

\bibitem[{Libovick{\`y} et~al.(2016)Libovick{\`y}, Helcl, Tlust{\`y}, Bojar,
  and Pecina}]{libovicky2016cuni}
Jind{\v{r}}ich Libovick{\`y}, Jind{\v{r}}ich Helcl, Marek Tlust{\`y},
  Ond{\v{r}}ej Bojar, and Pavel Pecina. 2016.
\newblock Cuni system for wmt16 automatic post-editing and multimodal
  translation tasks.
\newblock In \emph{Proceedings of the First Conference on Machine Translation:
  Volume 2, Shared Task Papers}, volume~2, pages 646--654.

\bibitem[{Manning et~al.(2014)Manning, Surdeanu, Bauer, Finkel, Bethard, and
  McClosky}]{coreNLP}
Christopher~D. Manning, Mihai Surdeanu, John Bauer, Jenny Finkel, Steven~J.
  Bethard, and David McClosky. 2014.
\newblock \href {http://www.aclweb.org/anthology/P/P14/P14-5010} {The
  {Stanford} {CoreNLP} natural language processing toolkit}.
\newblock In \emph{Association for Computational Linguistics (ACL) System
  Demonstrations}, pages 55--60.

\bibitem[{Mathews et~al.(2018)Mathews, Xie, and He}]{mathews2018semstyle}
Alexander Mathews, Lexing Xie, and Xuming He. 2018.
\newblock Semstyle: Learning to generate stylised image captions using
  unaligned text.
\newblock In \emph{The IEEE Conference on Computer Vision and Pattern
  Recognition (CVPR)}.

\bibitem[{Mitchell et~al.(2018)Mitchell, Ferraro, Misra
  et~al.}]{mitchell2018proceedings}
Margaret Mitchell, Francis Ferraro, Ishan Misra, et~al. 2018.
\newblock Proceedings of the first workshop on storytelling.
\newblock In \emph{Proceedings of the First Workshop on Storytelling}.

\bibitem[{Popovic et~al.(2014)Popovic, Lommel, Burchardt, Avramidis, and
  Uszkoreit}]{popovic2014relations}
Maja Popovic, Arle Lommel, Aljoscha Burchardt, Eleftherios Avramidis, and Hans
  Uszkoreit. 2014.
\newblock Relations between different types of post-editing operations,
  cognitive effort and temporal effort.
\newblock In \emph{Proceedings of the 17th Annual Conference of the European
  Association for Machine Translation (EAMT 14)}, pages 191--198.

\bibitem[{Purdy et~al.(2018)Purdy, Wang, He, and Riedl}]{purdy2018predicting}
Christopher Purdy, Xinyu Wang, Larry He, and Mark Riedl. 2018.
\newblock Predicting generated story quality with quantitative measures.
\newblock In \emph{Fourteenth Artificial Intelligence and Interactive Digital
  Entertainment Conference}.

\bibitem[{Roemmele and Gordon(2018{\natexlab{a}})}]{roemmele2018linguistic}
Melissa Roemmele and Andrew Gordon. 2018{\natexlab{a}}.
\newblock Linguistic features of helpfulness in automated support for creative
  writing.
\newblock In \emph{Proceedings of the First Workshop on Storytelling}, pages
  14--19.

\bibitem[{Roemmele and Gordon(2018{\natexlab{b}})}]{roemmele2018automated}
Melissa Roemmele and Andrew~S Gordon. 2018{\natexlab{b}}.
\newblock Automated assistance for creative writing with an rnn language model.
\newblock In \emph{Proceedings of the 23rd International Conference on
  Intelligent User Interfaces Companion}, page~21. ACM.

\bibitem[{Simard et~al.(2007)Simard, Goutte, and
  Isabelle}]{simard2007statistical}
Michel Simard, Cyril Goutte, and Pierre Isabelle. 2007.
\newblock Statistical phrase-based post-editing.
\newblock In \emph{Proceedings of NAACL HLT}, pages 508--515.

\bibitem[{Sutskever et~al.(2014)Sutskever, Vinyals, and
  Le}]{sutskever2014sequence}
Ilya Sutskever, Oriol Vinyals, and Quoc~V Le. 2014.
\newblock Sequence to sequence learning with neural networks.
\newblock In \emph{Advances in neural information processing systems}, pages
  3104--3112.

\bibitem[{Twain(1876)}]{twain1876adventures}
Mark Twain. 1876.
\newblock \emph{The Adventures of Tom Sawyer}.
\newblock American Publishing Company.

\bibitem[{Vaswani et~al.(2017)Vaswani, Shazeer, Parmar, Uszkoreit, Jones,
  Gomez, Kaiser, and Polosukhin}]{vaswani2017attention}
Ashish Vaswani, Noam Shazeer, Niki Parmar, Jakob Uszkoreit, Llion Jones,
  Aidan~N Gomez, {\L}ukasz Kaiser, and Illia Polosukhin. 2017.
\newblock Attention is all you need.
\newblock In \emph{Advances in Neural Information Processing Systems}, pages
  5998--6008.

\bibitem[{Wang et~al.(2018)Wang, Chen, Wang, and Wang}]{acl2018wang}
Xin Wang, Wenhu Chen, Yuan-Fang Wang, and William~Yang Wang. 2018.
\newblock No metrics are perfect: Adversarial reward learning for visual
  storytelling.
\newblock In \emph{Proceedings of the 56th Annual Meeting of the Association
  for Computational Linguistics}, Melbourne, Victoria, Australia. ACL.

\end{thebibliography}
\bibliographystyle{acl_natbib}

\end{document}